\documentclass{article}

\usepackage{arxiv}

\usepackage[utf8]{inputenc} % allow utf-8 input
\usepackage[T1]{fontenc}    % use 8-bit T1 fonts
\usepackage{hyperref}       % hyperlinks
\usepackage{booktabs}       % professional-quality tables
\usepackage{amsfonts}       % blackboard math symbols
\usepackage{nicefrac}       % compact symbols for 1/2, etc.
\usepackage{microtype}      % microtypography
\usepackage{graphicx}
\usepackage{graphicx}
\usepackage{natbib}
\usepackage{authblk}
\usepackage{url}
\usepackage{multirow}
\usepackage{amsmath}
\usepackage{amsfonts}
\usepackage{xspace}
\usepackage{subfigure}
\usepackage[skip=2pt]{caption}
\usepackage[flushleft]{threeparttable}
\usepackage{changepage}
\usepackage{setspace}
\graphicspath{ {./images/} }

\title{Diagnostic Captioning: A Survey}

\author[1,*]{\bf \large John Pavlopoulos}
\author[1,2]{\bf \large Vasiliki Kougia}
\author[2]{\bf \large Ion Androutsopoulos}
\author[3]{\bf \large Dimitris Papamichail}

\affil[1]{Department of Computer and Systems Sciences, Stockholm University,
Borgarfjordsgatan 12, 164 55 Kista, Sweden}
\affil[2]{Department of Informatics, Athens University of Economics and Business, Patission 76, GR-104 34, Athens, Greece}
\affil[3]{Nuclear Medicine Department, Medical Diagnostic Center Kosmoiatriki, Patission 237, GR-112 54, Athens, Greece}
\affil[*]{Corresponding author. Email: annis@aueb.gr}

\begin{document}
\maketitle

\newcommand\imageclef{\textsc{ImageCLEF}\xspace}
\newcommand\freq{\textsc{Frequency}\xspace}
\newcommand\knn{\textsc{NearestNeighbor}\xspace}
\newcommand\imageclefcaption{\textsc{ICLEFcaption}\xspace}
\newcommand\peirgross{\textsc{PEIR Gross}\xspace}
\newcommand\peirradio{\textsc{PEIR Radio}\xspace}
\newcommand\iuxray{\textsc{IU X-Ray}\xspace}
\newcommand\mimic{\textsc{MIMIC-CXR}\xspace}
\newcommand\dcaptioning{\emph{Diagnostic Captioning}\xspace}
\newcommand\captioning{\textsc{Captioning}\xspace}

\begin{abstract}
Diagnostic Captioning (DC) concerns the automatic generation of a diagnostic text from a set of medical images of a patient collected during an examination. DC can assist inexperienced physicians, reducing clinical errors. It can also help experienced physicians produce diagnostic reports faster. Following the advances of deep learning, especially in generic image captioning, DC has recently attracted more attention, leading to several systems and datasets. This article is an extensive overview of DC. It presents relevant datasets, evaluation measures, and up to date systems. It also highlights shortcomings that hinder DC's progress and proposes future directions.
\end{abstract}

% keywords can be removed

\section{Introduction} \label{sect:intro}
Medical Imaging is concerned with forming visual representations of the anatomy or a function of a human body using a variety of imaging modalities (e.g., X-rays, CT, MRI) \citep{suetens2009fundamentals,aerts2014decoding}. It is estimated that approximately one billion medical imaging examinations are performed worldwide annually \citep{krupinski2010current}, and the number of medical imaging examinations per year continues to rise in developed countries \citep{brady2017error}. More sophisticated medical imaging systems lead to more images per examination and radiologist \citep{chokshi2015diagnostic}. The total workload has increased by 26\% from 1998 to 2010 \citep{chokshi2015diagnostic}, and radiologists must now interpret more images during work time compared to similar examinations performed 10--20 years ago. Radiologists need to consider the examination's medical images, patient history, previous examinations, consult recent bibliography, and prepare a medical report. An increased workload in this demanding task increases the likelihood of medical errors (e.g., when radiologists are tired or pressured). These errors are not rare \citep{berlin2007accuracy} and they will also be present in the  medical report, which is what referring physicians (who ordered the examination) mostly consider. Consequently, tools that would help radiologists produce higher quality reports (e.g., without missing important findings or reporting wrong findings when they are inexperienced) in less time (e.g., by providing them with a draft report) could have a significant impact.

\begin{figure}[ht]
\centering
    \begin{minipage}{\textwidth}
    \subfigure[Generic image captioning]{
        \centering
        \includegraphics[width=0.45\textwidth]{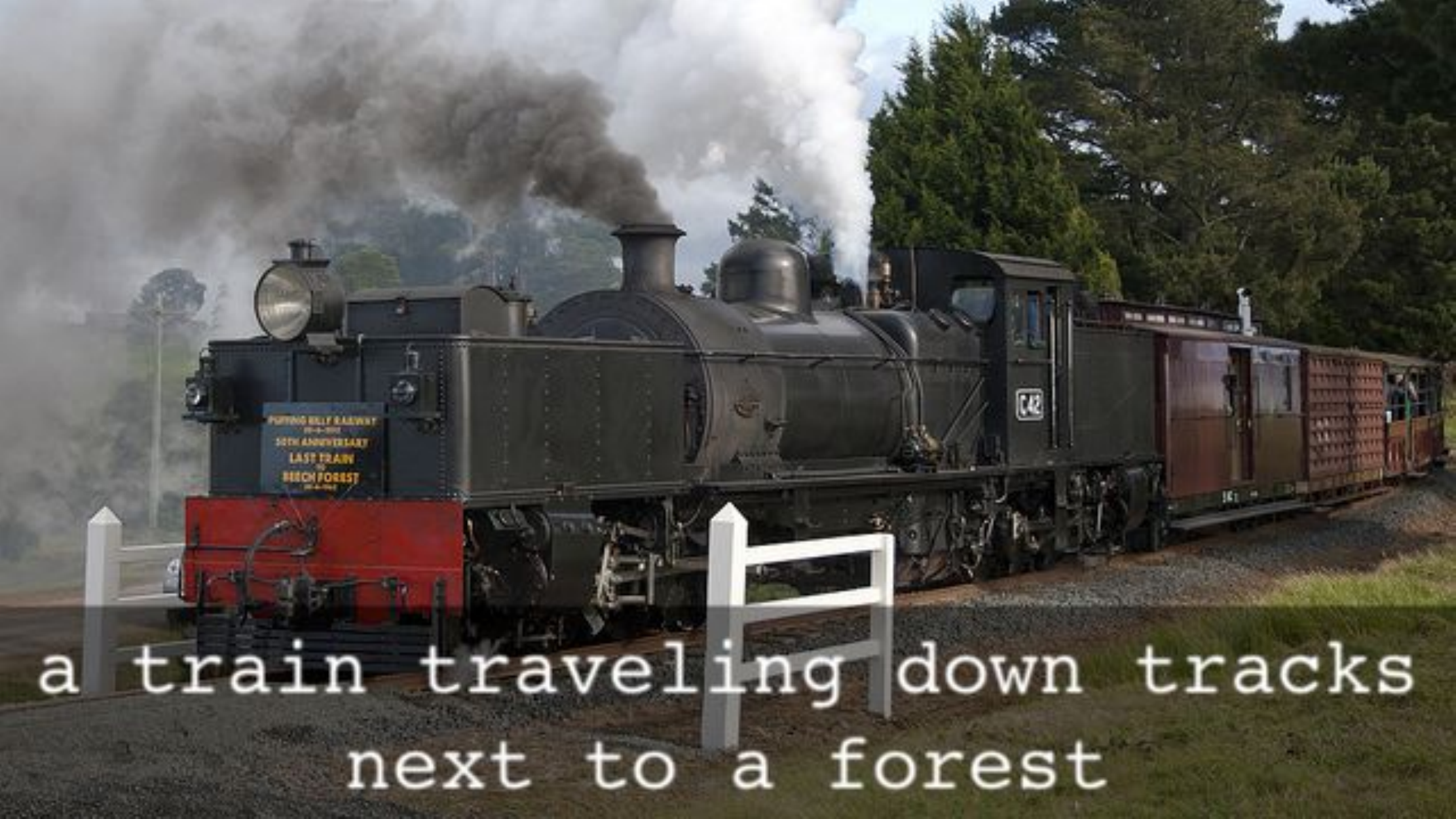}
        \label{fig:captioninga}
    }
    %\vspace{\floatsep}
    \subfigure[Diagnostic image captioning]{
        \centering
        \includegraphics[width=0.45\textwidth]{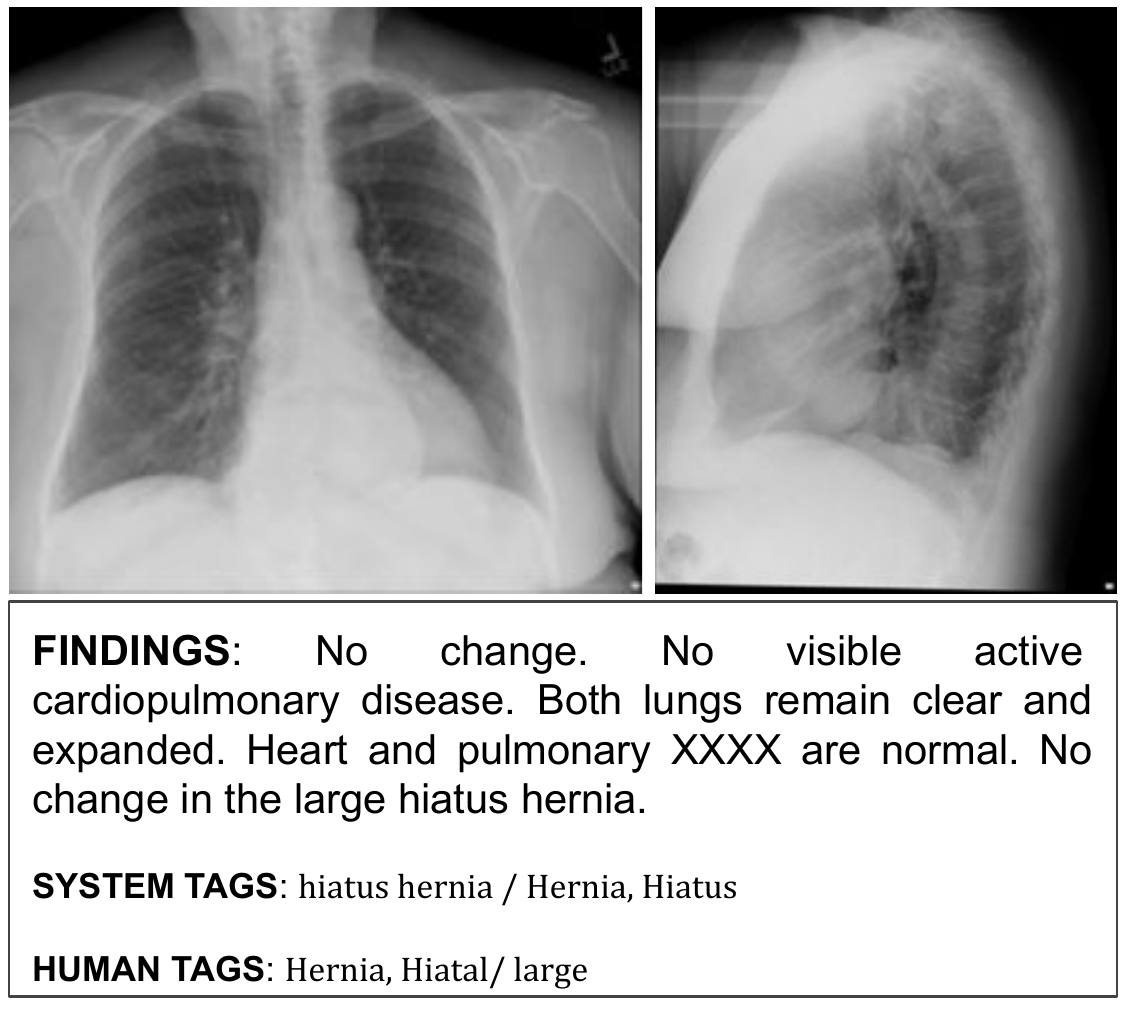}
        \label{fig:captioningb}
    }
    \end{minipage}
\caption{(a) Caption produced by the \emph{generic captioning} system of \citet{Vinyals2015}. (b) Two images from an X-ray examination along with: the corresponding human-authored `FINDINGS' section, from the IUXRay dataset (Section~
\ref{ssec:iuxray}); silver labels (tags) automatically extracted from the human-authored findings (`system tags'); and gold labels provided by physicians (`human tags'). The diagnostic caption refers to `no change' with respect to a previous examination, but there is no link to information and images of the previous examination in this particular dataset. The `XXXX' is due to a (presumably automatic) de-identification process.}
\end{figure}

\dcaptioning (DC) systems encode the medical images of a patient's examination (or `study') and generate a full or partial draft of the report. Fig.~\ref{fig:captioningb}, for example, shows only the `FINDINGS' section of the report. Although we are not aware of any clearly articulated description of the exact goals of research on DC systems, the main goals seem to be to (a) increase the \emph{throughput} of medical imaging departments \citep{hosny2018artificial}, (b) reduce \emph{medical errors} \citep{fazal2018past}, and (c) reduce the \emph{cost} of medical imaging examinations \citep{liew2018future}. Advances in image to text generation \citep{Bernardi2016}, especially deep learning methods \citep{Goodfellow2016,Charniak2018} for generic \emph{image captioning} (Fig.~\ref{fig:captioninga}) \citep{Vinyals2017,Sharma2018,Hossain2019,Liu2019survey,Donahue2015,Lu2017,Rennie2017}, have recently led to increased research interest in DC \citep{Jing2018,Li2019,Liu2019,Kougia2019}. However, despite its importance and recent popularity, DC still suffers from shortcomings in methods, datasets, and evaluation measures. This article attempts to review all the published work on DC, outlining the major problems, and proposing future research directions.

\smallskip
\noindent\textbf{DC methods} usually employ the encoder-decoder architecture \citep{Cho2014}, heavily ignoring retrieval-based approaches. In a similar manner, \citet{Monshi2020} recently surveyed deep learning based DC methods, but only considered systems where the diagnostic text is generated from scratch, using a Recurrent Neural Network (RNN) decoder. However, recent studies show that retrieval-based approaches, where text from similar previous exams is reused, are very competitive despite their simplicity, often surpassing much more complex methods \citep{Kougia2019}. Even the simple nearest neighbor approach, where the diagnosis of the visually most similar study is retrieved and reused, has been reported to outperform all other approaches in clinical recall \citep{Liu2019}. This could be due to the fact that retrieval can assist in capturing  factual knowledge \citep{khandelwal2019} and negation; the latter is particularly important in clinical text \citep{Kassner2020}. Or it could be due to the fact that medical reports seem to use a template-based language, where reports of the same findings are almost identical across different patients.
We provide a segmentation of current DC systems by the kinds of methods they use, also reporting evaluation scores for each system with all available measures.

\smallskip
\noindent\textbf{Datasets} that have been used in previous DC work \citep{johnson2019,DF2015,Wang2018,Gale2018,Jing2018,Li2018} are not all publicly available. Out of the four publicly available ones, \peirgross\footnote{PEIR Gross is a subcollection of the PEIR Digital Library (\url{https://peir.path.uab.edu/library/}).} and \imageclef \citep{ImageCLEF2018} suffer from severe shortcomings \citep{Kougia2019}, which are also discussed briefly below. Hence, in this work we focus mostly on studying and discussing the characteristics of the remaining two datasets, namely \iuxray \citep{DF2015} and \mimic \citep{johnson2019}. Interestingly, previous research does not always use the same parts of the medical reports of these datasets
(some use the `FINDINGS' section only, others include the `IMPRESSION'), and most previous articles do no use a common training-development-test split of the data.
For \iuxray, which is the most commonly used dataset, we use the split of \citet{Li2018}, who recently used it to evaluate multiple DC systems. We also release (in supplementary material that will accompany the camera-ready) instructions on how to obtain and use this split.

\smallskip
\noindent\textbf{Evaluation measures} employed by previous DC research mainly
assess lexical overlap between machine-generated and human-authored captions \citep{Kougia2019}, without directly assessing clinical correctness. This can lead to cases where a clinically wrong generated report can be scored higher than a clinically correct one \citep{Zhang2019}. Current methods for automatically measuring clinical correctness produce results of poor quality, because: (a) they only consider the presence (or absence) of particular medical terms in the reports \citep{Li2019}, for example `pneumothorax' would be considered a positive find in `no pneumothorax is observed' \citep{Zhang2019}; or (b) they rely on the responses of rule-based automatic annotators \citep{Liu2019}, for example to obtain the `system tags' in  Fig.~\ref{fig:captioningb}, whose accuracy cannot be guaranteed; or (c) they use crowd workers \citep{Li2018}, who are not necessarily medical experts or trained in medical informatics.

\smallskip
Excluding an earlier version of this survey \citep{Kougia2019}, the only other DC survey we are aware of is that of \citet{Monshi2020}.\footnote{The earlier version of this survey \citep{Kougia2019} did not consider the recent \mimic dataset, did not investigate as thoroughly the shortcomings of existing DC datasets and evaluation measures, and considered fewer (and less recent) DC methods, providing less information about them.} As already pointed out, the latter considers only DC methods that generate diagnostic text from scratch using an RNN decoder, whereas we also consider retrieval-based methods, which are often very competitive in DC. We also scrutinize much more the datasets we consider. For instance, we study how often the diagnostic reports are very similar across patients, the class imbalance between reports with no findings vs.\ reports that report abnormalities, or to what extent relevant information is missing (e.g., reports referring to images that have been removed during anonymization, or sections that require access to unavailable previous examinations of the same patient). Furthermore, we provide a more extensive discussion of evaluation measures; for example, we also cover clinical correctness measures, apart from word overlap measures that \citet{Monshi2020} mostly focus on; and we demonstrate the shortcomings of current evaluation measures using concrete DC examples. A final difference from the survey of \citet{Monshi2020} is that we assume the reader is familiar with commonly used machine learning algorithms, including currently widely used deep learning (DL) models like Convolutional and Recurrent neural networks (CNNs, RNNs). This allows us to present and compare DC methods more succinctly. Readers who lack this background should consult introductory machine learning and DL textbooks first \citep{Murphy2012,Goodfellow2016,Goldberg2017,Charniak2018}.

%%%%%%%%%%%%%%%%%%%%%%%%%%%%%%%%%

\section{Diagnostic Captioning Datasets} 
\label{sect:data}

Datasets for DC comprise medical images and associated diagnostic reports. In previous work \citep{Kougia2019}, we reported that three publicly available datasets can be used for DC research, namely \peirgross, \imageclefcaption \citep{ImageCLEF2018}, and \iuxray \citep{DF2015}. We concluded, however, that the first two datasets suffer from severe shortcomings. 
% e.g., 
Most importantly, they contain photographs and captions from the figures of scientific articles, instead of real diagnostic medical images and reports; hence, they are inappropriate for realistic DC research. The third dataset, \iuxray, which contains X-ray images and medical reports, is appropriate if we ignore its small size. In our previous work \citep{Kougia2019}, we did not consider \mimic \citep{johnson2019}, a fourth and the largest to date publicly available DC dataset, which was released later, and we only partially explored \iuxray. In this work, we focus on the latter two quality datasets, \mimic and \iuxray, referring readers interested in \imageclef and \peirgross to our previous work \citep{Kougia2019}. Datasets that do not comprise both medical images and diagnostic reports, or that are not publicly available, are not considered further in this study. Such datasets are BCIDR \citep{Zhang2017b}, consisting of 1,000 pathological bladder cancer images, each with five reports, which is not publicly available; Frontal Pelvic X-Rays \citep{Gale2018}, which comprises 50,363 images, each accompanied by a radiology report simplified to follow a standard template, but is not publicly available; and Chest X-Ray 14 \citep{Wang2018}, which is publicly available, but does not include any medical reports in its public version. Results on these datasets are included, however, in Table~\ref{tab:results} for completeness.

\begin{figure}[ht]
\centering
    \begin{minipage}{\textwidth}
    \subfigure[\iuxray]{
        \centering
        \includegraphics[width=.45\textwidth]{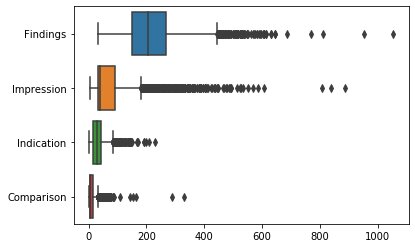}
    }
    \vspace{\floatsep}
    \subfigure[\mimic]{
        \centering
        \includegraphics[width=.45\textwidth]{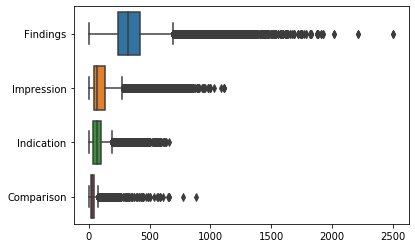}
    }
    \end{minipage}
\caption{Boxplots of section lengths in characters, in (a) \iuxray and (b) \mimic.}
\label{fig:data_stats}
\end{figure}

Radiologists usually document their findings in titled sections, following standardized document structure templates. However, the sections found in the reports of \iuxray are not always the same as the sections found in the reports of \mimic.  `FINDINGS', `COMPARISON', `INDICATION', `IMPRESSION' are sections found in both datasets, among which `FINDINGS' and `IMPRESSION' are of primary interest \citep{johnson2019}. The `FINDINGS' section, which is usually the lengthier one (Fig.~\ref{fig:data_stats}), describes the imaging characteristics of a body structure of function that can have a clinical impact. `COMPARISON' contains previous information about the patient, often from preceding medical exams, but never the whole report or the medical images of the previous exams. This means that it is almost impossible even for a radiologist to generate this section without the previous referred exams. The same applies to the `INDICATION' section, which conveys the medical reason for the patient to be subjected to the examination (e.g., symptoms). Hence, the `COMPARISON' and `INDICATION' sections cannot be generated by DC methods. Instead, they could be treated as given and they could, at least in principle, assist the process of generating the `FINDINGS', although current DC methods attempt to generate the `FINDINGS' directly from the images, without consulting the `COMPARISON' and `INDICATION' sections. 
All the aforementioned sections could in turn assist the process of generating the final section `IMPRESSION', although again current DC methods try to generate the `IMPRESSION' directly from the images.
The `IMPRESSION' usually summarizes the most important findings and interprets their clinical value, giving the referring physician a direction for the management of the disease or a final diagnosis. 
However, sometimes the `IMPRESSION' (or `FINDINGS') includes a conclusion that does not follow from the previous sections and the images of the current exam. For example, a conclusion may be the result of comparing the current exam with a previous one. Unfortunately, the dataset may omit the previous exam(s), as in \iuxray; or it may hide the dates and times of the exams of each patient, as in \mimic, making it impossible to identify the previous exams.

\begin{table}
\centering
\begin{threeparttable}
    \small
    \caption{Frequent `FINDINGS' sections with no abnormality reported in \iuxray. Diagnostic findings with no abnormality reported (no abnormality tag assigned by experts) follow few templates. In \iuxray, for example, the top three most frequent `FINDINGS' sections occur (exactly the same) in 148 reports (studies). By contrast, reports of abnormalities are less standardized.}
    \begin{tabular}{c p{12cm}} 
    \hline
         \bf Studies (\#) & \bf Reported Findings \\\hline
         51 & The heart and lungs have XXXX XXXX in the interval. Both lungs are clear and expanded. Heart and mediastinum normal.\\ \hline
         51 & The heart is normal in size. The mediastinum is unremarkable. The lungs are clear.\\ \hline
         46 & Heart size normal. Lungs are clear. XXXX are normal. No pneumonia, effusions, edema, pneumothorax, adenopathy, nodules or masses.\\\hline
    \end{tabular}
    \label{tab:normal_findings}
\end{threeparttable}
\end{table}

\begin{table}
    \centering
    \begin{threeparttable}
    \caption{Number of reports per dataset (1st row) and of reports whose diagnostic text includes a section with Findings, Impression, Indication or Comparison (2nd to 5th row).}
        %\large
            \begin{tabular}{p{9cm} c c}
            \hline \bf Number of Reports & \bf \iuxray & \bf \mimic \\ \hline
            \sc \# Reports & 3851 (100\%) & 227,835 (100\%) \\ \hline
            \sc \# Reports with Findings & 3337 (87\%) & 149,758 (66\%) \\
            \sc \# Reports with Impression & 3851 (99\%) & 187,793 (82\%) \\
            \sc \# Reports with Indication & 3765 (98\%) & 155,607 (68\%) \\
            \sc \# Reports with Comparison & 3252 (84\%) & 154,586 (68\%) \\
            \hline
            \end{tabular}
    \label{tab:data_stats}
    \end{threeparttable}
\end{table}

The majority of the reports in both \iuxray and \mimic concern cases where there is no disease or abnormality. In these cases, the diagnostic text is often very similar or identical across different exams (Table~\ref{tab:normal_findings}). The section that exists most often in both datasets is `IMPRESSION' (Table~\ref{tab:data_stats}). However, it is also the section with the shortest text on average (Fig.~\ref{fig:data_stats}) and often comprises conclusions drawn from information not included in the datasets, as already noted. Some previous work used only the `IMPRESSION' section as the target text to be generated \citep{Shin2016a}, but most previous work either uses the `FINDINGS' as the target \citep{Liu2019,Li2018} or aims to generate the concatenation of the two sections \citep{Shin2016a,Jing2018}.

Another publicly available dataset is PadChest \citep{Bustos2020}, which comprises 160,868 chest X-Rays from 69,882 patients. However, its diagnostic texts (in Spanish) are not complete reports or complete sections of reports, but text snippets (not necessarily well formed sentences or paragraphs) that were extracted from the reports with regular expressions. Therefore, we exclude this dataset from our study, since it does not contain texts of the kinds the DC methods we consider aim to generate (entire reports or particular sections of reports).

\subsection{\iuxray}
\label{ssec:iuxray}

\citet{DF2015} created a dataset of radiology examinations, comprising X-ray images and reports authored by radiologists. They publicly released an anonymized version of the dataset through the Open Access Biomedical Image Search Engine (OpenI).\footnote{\url{https://openi.nlm.nih.gov/}} The dataset consists of 3,955 reports, one per patient, all in English, and 7,470 DICOM images.\footnote{Each report is an XML file and images can be also downloaded in the PNG format. DICOM is a standard for medical images; see \url{https://www.dicomstandard.org/current/}.} We found that 3,851 reports (97.4\%) are linked to at least one image and are thus valid for our study. Among the 3,851 reports, 599 (15.6\%) do not include a `COMPARISON' section, 86 (2.2\%) do not include the `INDICATION' section, 31 (0.1\%) do not include `IMPRESSION', while 514 (13.3\%) do not include `FINDINGS'. Among the 3,337 reports that comprise both a `FINDINGS' section and at least one image, 
only 2,553 (76.5\%) have a unique `FINDINGS' section, i.e., the text of their `FINDINGS' is not the same in any other report. The `FINDINGS' section of the remaining 23.5\% reports is exactly the same in two or more other reports, and in these cases the reports describe mainly normal findings. For example, the most frequent `FINDINGS' text (in 51 reports) is: \emph{``The heart is normal in size. The mediastinum is unremarkable. The lungs are clear.''} The 10 most frequent `FINDINGS' sections, all describing normal findings, occur in 344 reports in total, which is 10.3\% of all the reports and 43.9\% of the non-unique `FINDINGS'. This gives an advantage to retrieval-based approaches, which have been reported to achieve surprisingly high performance overall \citep{Kougia2019,Liu2019,Boag2020}.

\citet{DF2015} initially collected and de-identified 4k examinations from two hospitals, 2k from each one. They used software to de-identify the texts \citep{Friedlin2008}, and the Clinical Trial Processor to de-identify the images.\footnote{\url{mirc.rsna.org/download/CTP-installer.jar}} They reported excellent de-identification results (no sensitive information found by human annotators), but we found some cases where the software had damaged the diagnostic text. For example, \emph{``Cardiomediastinal silhouette is XXXX''} is not informative and we cannot accurately infer the finding. Demner-Fushman et al.\ then discarded four exams that either did not comprise both a lateral and a posterioranterior chest image, or had diagnostic text that did not include clearly separated sections for the findings and the impression (or the diagnosis). They also discarded 41 exams containing information that could reveal the patient identities. Two of them revealed information such as address and dates, which the Health Insurance Portability and Accountability Act of 1996 (HIPAA) considers identifiers that may reveal personal information. The remaining 39 were exams with images that comprised teeth, partial jaw, jewelry, or partial skull. Although not explicitly flagged as sensitive by HIPAA, these images might identify the patients. These 39 exams had all of their images in this category and were completely removed from the dataset. However, 432 (10.9\%) other exams had at least one image removed for the same reasons, but were apparently retained in the dataset (without the removed images). These 432 cases are problematic for training and evaluation purposes, because the gold (target) diagnostic text that the systems are required to generate may refer to an absent image. 

The `IMPRESSION' and `FINDINGS' sections were used by \citet{DF2015} to manually associate each report with a number of tags, shown as `human tags' in Fig.~\ref{fig:captioningb}. Two human annotators were used, both trained in medical informatics. The tags were MeSH terms, supplemented with Radiology Lexicon terms (RadLex).\footnote{Consult \url{https://www.nlm.nih.gov/mesh/meshhome.html} and \url{http://radlex.org/}.} Each annotation label (term) referred to a pathology, a foreign body or transplant, anatomy (human body parts), signs (imaging observations), or attributes (object or disease characteristics). The annotators were instructed not to assign labels for negated terms (e.g., `no signs of tuberculosis') or inconclusive findings (e.g., introduced by `possibly' or `maybe', but not `probably' or `likely') of the `IMPRESSION' and `FINDINGS'.\footnote{The annotation guidelines are publicly available \citep{DF2015}.} In addition to the human tags, each report was associated with tags automatically extracted from the `IMPRESSION' and `FINDINGS' by Medical Text Indexer; the resulting tags are called `MTI encoding' and they are shown as `system tags' in Fig.~\ref{fig:captioningb}. As shown in the example of Fig.~\ref{fig:captioningb}, the system tags are single words or terms (e.g., `Hiatus'), while human tags follow a different pattern, which may combine anatomical site and type (e.g., `Hiatal/large'). Surprisingly, although the dataset comprises both human and system tags, only the latter have been used to evaluate (or train) DC systems so far; we discuss evaluation measures that use tags  in Section~\ref{ssec:clinical_measures} below.

\subsection{\mimic}
\label{ssec:mimic}
\mimic comprises 377,110 chest X-rays associated with 227,835 medical reports, from 64,588 patients of the Beth Israel Deaconess Medical Center examined between 2011 and 2016.\footnote{When \mimic was first introduced, these numbers were reported to be slightly different \citep{johnson2019}. The numbers we provide refer to the v2.0.0.~version that can be found at \url{https://mimic-cxr.mit.edu/}}
The reports were de-identified to satisfy HIPAA requirements (Section~\ref{ssec:iuxray}). The images are chest radiographs, obtained from the hospital's Picture Archiving and Communication System (PACS) in DICOM format. To remove all Protected Health Information (PHI), the images were processed to remove annotations imprinted in them (e.g., image orientation, anatomical position of the subject, timestamp of image capture). A custom algorithm was used for this purpose, based on image pre-processing and optical character recognition to detect text; all pixels within a bounding box containing this information were set to black. Two independent reviewers were employed upon de-identification to examine 6,900 radiographs for PHI, and they found no PHI.

The reports of \mimic are written in English and their text is separated in sections, following document structure templates.
Unlike \iuxray, where the boundaries of sections are made explicit by the XML markup, the section boundaries of \mimic reports are not explicitly marked up. However, the section headings of \mimic are written in upper case, followed by a column (e.g., ``FINDINGS:''). Apart from the sections described in the discussion of \iuxray (Section~\ref{ssec:iuxray}), some reports of \mimic include other sections, such as `HISTORY', `EXAMINATION', or `TECHNIQUE', but not in a consistent manner, because the structure of the reports and section names were not enforced by the hospital's user interface \citep{johnson2019}.

\section{Evaluation Measures for Diagnostic Captioning}
\label{sect:eval}

\begin{table}[]
\centering \small
\begin{threeparttable}
    \setlength\tabcolsep{4pt}
    \caption{Using the BLEU (B1, B2, B3, B4), METEOR (Met), and ROUGE (Rou) automatic evaluation measures to score clinically correct ($H_{cc}$) and clinically incorrect ($H_{ci}$) hypothetical diagnoses that paraphrase three reference (gold, human) diagnoses ($R^a, R^b, R^c$). Percentage scores are reported. The first triplet ($R^a$, $H^a_{cc}$, $H^a_{ci}$) is from \citet{Zhang2019}, while the rest were generated by a medical expert during our work. The last two columns show Precision (Pre) and Recall (Rec) scores of the `system' tags of each $H_{cc}$ and $H_{ci}$, which were extracted by the CheXpert labeller. All six automatic measures (B1, B2, B3, B4, Met, Rou) score the clinically incorrect diagnoses ($H_{ci}$) much higher than the clinically correct ones ($H_{cc}$).}
    \begin{tabular}{ c p{3.5cm} c c c c c c p{2.2cm} c c}
        \hline
         \sc Type & \sc Clinical diagnosis
         & \sc B1 & \sc B2 & \sc B3 & \sc B4 & \sc Met & \sc Rou & \sc Tags & \sc Pre & \sc Rec \\\hline \hline

         $R^{a}$ & Pneumothorax is seen. Bilateral pleural effusions continue. 
         &&&&&&& 
         Pneumothorax, Pleural Effusion
         &&\\ \hline
         $H^{a}_{cc}$ & Pneumothorax is observed on radiograph. Bilateral pleural effusions continue to be seen.
         & 58.3 & 46.1 & 34.9 & 26.2 & 46.1 & 66.3 &
         Pneumothorax, Pleural Effusion
         &100.0&100.0\\ \hline
         $H^{a}_{ci}$ & No pneumothorax is observed. Bilateral pleural effusions continue.
         & \bf 75.0 & \bf 65.5 & \bf 52.3 & \bf 41.1 & \bf 55.2 & \bf 81.0 &
         Pleural Effusion
         &100.0&50.0\\\hline \hline
         
         $R^b$ &  Stable cardiomegaly and mild bilateral interstitial opacities which represent mild pulmonary edema.
         &&&&&&&
         Cardiomegaly, Lung Opacity, Edema
         &&\\ \hline
         $H^{b}_{cc}$ & Enlarged heart without notable variation and mild densities in both lungs, compatible with fluid accumulation.
         & 13.3 & 9.8 & 0.0 & 0.0 & 5.8 & 15.1 &
         Cardiomegaly
         &100.0&33.3\\ \hline
         $H^{b}_{ci}$ & Decreased cardiomegaly without considerable bilateral interstitial opacities, which exclude acute pulmonary edema.
         & \bf 58.3 & \bf 46.1 & \bf 34.9 & \bf 26.2 & \bf 29.0 & \bf 58.3 & 
         Cardiomegaly
         &100.0&33.3\\\hline \hline
         
         $R^c$ &  Mildly prominent perihilar opacities, due to bronchovascular crowding.
         &&&&&&&
         Lung Opacity
         &&\\ \hline
         $H^{c}_{cc}$ & Bilateral interstitial densities in the hilar region, caused by increased bronchovascular markings.
         & 8.3 & 0.0 & 0.0 & 0.0 & 10.4 & 10.4 &
         \sc No Finding
         &0&0\\ \hline
         $H^{c}_{ci}$ & There is no evidence of prominent perihilar opacities or bronchovascular crowding.
         & \bf 45.5 & \bf 36.9 & \bf 24.7 & \bf 0.0 & \bf 33.9 & \bf 54.2 & 
         \sc No Finding
         &0&0\\\hline \hline
    \end{tabular}
    \label{tab:eval_fail}
    \end{threeparttable}
\end{table}

Texts generated by DC systems have so far been assessed mostly via automatic evaluation measures originating from machine translation and text summarization \citep{papineni2002,Lin2004,Anderson2016,Vedantam2015}, which, roughly speaking, count how many words or $n$-grams (phrases of $n$ consecutive words) are shared between the generated text and reference gold texts (typically human-authored). Such measures have been reported to correlate well with human judgments of information content (e.g., the degree to which the most important information is preserved in summaries) when the goal is to rank systems and when there are multiple gold references per generated text. However, as can be seen in Table~\ref{tab:eval_fail}, measures of this kind do not necessarily capture clinical correctness. 
$H^{a}_{ci}$ for example, which incorrectly reports `No pneumothorax', receives higher scores than $H^{a}_{cc}$, which correctly reports pneumothorax.
When the gold and the system-generated captions are automatically labelled for mentions of diseases or abnormalities \citep{Xue2018,Huang2019,Liu2019}, standard classification evaluation measures, such as Accuracy, Precision and Recall, can be applied. Automated labelling, however, can be inaccurate leading to mistaken results. In the second (b) and third (c) cases of Table~\ref{tab:eval_fail}, for example, the automatically assigned tags are the same for the correct ($H_{cc}$) and incorrect ($H_{ci}$) diagnoses.

In an interesting evaluation approach, \citet{Li2019} employed crowd-sourcing for 100 randomly selected studies (examinations). For each study, the annotators were shown three reports, produced by a physician, a baseline DC system, and a more elaborate DC system, respectively. Each annotator had to consult the report of the physician and choose the best system-generated report, based on the criteria of clinical \emph{correctness} of the reported abnormalities,
\emph{fluency}, and content \emph{coverage} compared to the ground truth report.
Although this approach is interesting in general, because it employs manual evaluation, the particular experiment of \citet{Li2019} raises doubts for three reasons. First, the medical background of the annotators was not reported and it may have been inadequate. Second, cases were excluded from the evaluation when no system-generated report was better than the other one, but these cases could be very frequent (e.g., if no system was good enough). Third, only 100 test studies were used and no statistical significance was reported. Consequently, we do not discuss the details and scores of this evaluation further.

\subsection{Word Overlap Measures}
\label{ssec:overlap_measure}

The most common word overlap measures are BLEU \citep{papineni2002}, ROUGE \citep{Lin2004}, and METEOR \citep{Banerjee2005}, which originate from machine translation and summarization, as already noted. The more recent CIDEr measure \citep{Vedantam2015}, which was designed for \emph{generic} (not medical) image captioning \citep{Kilickaya2016}, has been used in only two DC works so far \citep{Zhang2017b,Jing2018}. SPICE \citep{Anderson2016}, also designed for generic captioning \citep{Kilickaya2016}, has not been used at all in DC so far. We note again that all these are word overlap measures, which do not always capture clinical correctness \citep{Li2018}, as already discussed. This was also discussed by \citet{Zhang2019}, who used ROUGE to compare two medical statements, a clinically correct and a clinically incorrect one. Since the latter had more common words with the gold statement, it obtained a higher score. 
We included the example of Zhang et al.\ in Table~\ref{tab:eval_fail}, where we also used BLEU, METEOR, and ROUGE, along with Precision and Recall computed on CheXpert labels.

BLEU \citep{papineni2002} is the most common and the oldest among the word overlap measures that have been used in DC. It measures the word n-gram overlap between a generated and a ground truth caption. As this is a precision based measure, a brevity penalty is added to penalize short generated captions. BLEU-1 considers unigrams (i.e., single words), while BLEU-2, -3, -4 consider bigrams, trigrams and 4-grams respectively. The average of the four variants was used as the official measure in \imageclefcaption \citep{ImageCLEF2017,ImageCLEF2018}. 
METEOR \citep{Banerjee2005} extended BLEU-1 by employing the harmonic mean between Precision and Recall i.e., the $F_{\beta}$ score, but biased towards Recall ($\beta >1$). METEOR also employs Porter's stemmer and WordNet \citep{Fellbaum2012}, the latter to take synonyms into account.\footnote{\url{https://tartarus.org/martin/PorterStemmer/}} 
The $F_{\beta}$ score is then penalized up to 50\% when no common $n$-grams exist between the machine-generated description and the reference human description.
ROUGE-L \citep{Lin2014} is the ratio of the length of the longest common $n$-gram shared by the machine-generated description and the reference human description, to the size of the reference description (ROUGE-L Recall); or to the generated description (ROUGE-L Precision); or it is a combination of the two (ROUGE-L F-measure). We note that many ROUGE variants exist \citep{Graham2015}, based on different $n$-gram lengths, stemming, stopword removal, etc., but ROUGE-L is the most commonly used version in DC so far.
CIDER \citep{Vedantam2015} measures the cosine similarity between $n$-gram TF-IDF \citep{Manning2008} representations of the two captions; words are also stemmed. Cosine similarities are calculated for unigrams to 4-grams and their average is returned as the final evaluation score. The intuition behind using TF-IDF is to reward terms that are frequent in a particular caption being evaluated, while penalizing terms that are common across captions (e.g., stopwords). However, DC datasets have constrained vocabularies, and a common disease name may thus be mistakenly penalized. More importantly, CIDEr scores exceeding 100\% have been reported \citep{Liu2019}, which contradicts the measure's theoretic design. By using the official evaluation server implementation (CIDEr-D) \citep{Capeval2015}, we found cases where scores exceeding 100\% were indeed produced, which means that further investigation is required to check the correctness of the particular implementation of the measure and allow a fair comparison among systems. 
SPICE \citep{Anderson2016} extracts sets of tuples from the two captions (human and machine-generated), containing objects, attributes, and/or relations; e.g., \{(patient), (has, pain), (male, patient)\}. Precision and recall are computed between the two sets of tuples, also taking WordNet synonyms into account, and the $F_1$ score is returned. The authors of SPICE report improved results over both METEOR and CIDEr, but it has been noted that results depend on the quality of syntactic parsing \citep{Kilickaya2016}. When experimenting with an implementation of this measure,\footnote{\url{https://goo.gl/bo11Bz}} we noticed that long texts were not parsed at times and thus were not evaluated properly.

\subsection{Clinical Correctness Measures}
\label{ssec:clinical_measures}

The word overlap measures discussed above do not always capture clinical correctness, as already demonstrated. To overcome this problem, recent work has proposed new evaluation approaches based on classification evaluation measures, an approach we have already mentioned and we now discuss further. The clinical correctness of a generated caption is measured through a set of medical terms extracted from that caption (see Table~\ref{tab:eval_fail}). These terms are then compared to the ones from the gold caption, which may have been generated by humans (gold labels, as in \iuxray) or a system (silver labels), like the Medical Text Indexer \citep{Mork2013} or the CheXpert labeller \citep{Irvin2019}. In Table~\ref{tab:eval_fail} for example, CheXpert was used to annotate three reference diagnoses ($R_a, R_b, R_c$) along with their alternative correct and incorrect hypothetical diagnoses ($H_{cc}$, $H_{ci}$). In the topmost example, \textsc{Pleural Effusion} and \textsc{Pneumonothorax} have been correctly generated by CheXpert,\footnote{We followed the work of \citet{Liu2019}, who flipped a coin when CheXpert returned the `unsure' label.} for the reference diagnosis and the correct hypothetical diagnosis $H^{a}_{cc}$. For the incorrect $H^{a}_{ci}$, only \textsc{pleural effusion} was generated, leading to a perfect Precision (number of correctly assigned tags to the total number of assigned tags) and a reasonable 50\% Recall (number of correctly assigned tags to the number of gold tags). In the next example, CheXpert does not detect 2 out of 3 tags for $H^{b}_{cc}$, leading to a low 33.3\% Recall. 
In the lowermost example, where the reference was labelled with \textsc{lung opacity}, no tags were detected by CheXpert for the correct hypothetical diagnosis $H^{c}_{cc}$. This leads to zero Recall and undefined Precision (though Precision is often taken to be also zero in such cases). Interestingly, however, the incorrect $H^{c}_{ci}$, which has the same (equally bad) Precision and Recall as $H_{cc}$, got high scores in many word overlap measures in Table~\ref{tab:eval_fail}, showing a weakness of such measures with respect to clinical accuracy assessment.

\citet{Xue2018} were the first to use an evaluation measure that considers medical tags extracted from system-generated and human-authored reports. The authors called the measure Keyword Accuracy, but it should not be confused with the conventional classification Accuracy, since it only measures Recall. The authors, who used the \iuxray dataset for their study, compiled a list of tags per examination and  used it as the ground truth; the list consisted of the system-generated (MTI) tags and some of the human tags available in \iuxray. However, Xue et al.\ did not provide any further details (e.g., about the human tag selection criteria or how system and human tags were merged). 
\citet{Huang2019} followed the same approach, but they used only the MTI tags as their ground truth. In both of these works, however, where gold tags were compared with predicted tags, it is unclear how the predicted tags were extracted from the system-generated reports. \citet{Liu2019} used the CheXpert medical abnormality mention detection system \citep{Irvin2019}, which generates one out of 4 labels (presence, absence, negative, not sure) for each one of 14 thoracic diseases. 
For any given report, any disease for which the assigned label was `presence' was considered to be mentioned in that report. When the assigned label for a disease was `not sure', then the authors considered that the disease was mentioned in the report with 0.5 probability. This process was applied to both system-generated and gold reports and then micro- and macro-averaged Precision and Recall were computed (along with macro-averaged Accuracy).
A disadvantage of this approach is that it uses only 14 diseases, which is a very small number compared to the hundreds of abnormality tags of other works and the much wider variety of medical conditions physicians need to consider in practice. However, the work of \citet{Liu2019} can be used to highlight the limitations of Accuracy compared to Precision and Recall, when used to assess DC systems. A majority classifier (a system that always reported no findings, the majority prediction) obtained a higher Accuracy than more elaborate methods in the experiments of Liu et al. More generally, it is well known in Machine Learning that a large class imbalance may lead to misleadingly high Accuracy, which is why Precision, Recall, and F1 are used instead in such cases. In Table~\ref{tab:results}, we computed and report the harmonic mean (F1) of Precision and Recall using the results of \citet{Liu2019}. We note that Receiver Operating Characteristics (ROC) curves, which plot the true positive rate against the false positive rate for different classification probability cut-offs, can also be used to get a better view of the performance of systems and baselines \citep{Swets1988}. Precision-Recall (PR) curves can be used in a similar manner. In both cases, the Area Under each Curve (AUC) can serve as a single evaluation score that aggregates results over different cut-offs. However, previous DC work does not provide enough information to reconstruct ROC or PR curves, and does not report AUC scores.

\section{Diagnostic Captioning Methods}
\label{sect:methods}

We now discuss the main types of DC methods, including their relation to generic image captioning. We also briefly cover early approaches that did not process medical images directly, but were fed with findings manually extracted from medical images, or that did not generate text, but were intended to help in the manual preparation of diagnostic reports.

\subsection*{Early approaches}

\citet{Varges2012} followed an ontology-based natural language generation approach to assist medical professionals turn cardiological findings (from diagnostic images) into %fluent 
readable
and informative textual descriptions. The input to the text generator, however, was not directly a medical image, but triplets like $<$\textit{right atrium}, \textit{size}, \textit{normal}$>$. 
From a different perspective, \citet{Schlegl2015} used medical images and their diagnostic reports as input to a Convolutional Neural Network (CNN) to classify voxels (3D pixels) as intraretinal cystoid fluid, subretinal fluid, or normal retinal tissue, 
with the help of concepts automatically extracted from the text of the corresponding report; in this case, the report was part of the system's input, not the system's output. \citet{Kisilev2015a,Kisilev2015b}
performed semi-automatic lesion detection and contour extraction from medical images. Structured Support Vector Machines \citep{Tsochantaridis2004} were then used to generate semantic tags, originating from a radiology lexicon, for each lesion. In later work, \citet{Kisilev2016} used a CNN to detect Regions of Interest (ROIs) in the images, then fully connected layers to assign predefined features describing abnormalities to each ROI. The assigned features were finally filled into sentence templates to generate captions. We discuss template-based text generation below.

\subsection*{Generic Image Captioning vs.\ Diagnostic Captioning} 

\begin{figure}
    \centering
    \includegraphics[width=\textwidth]{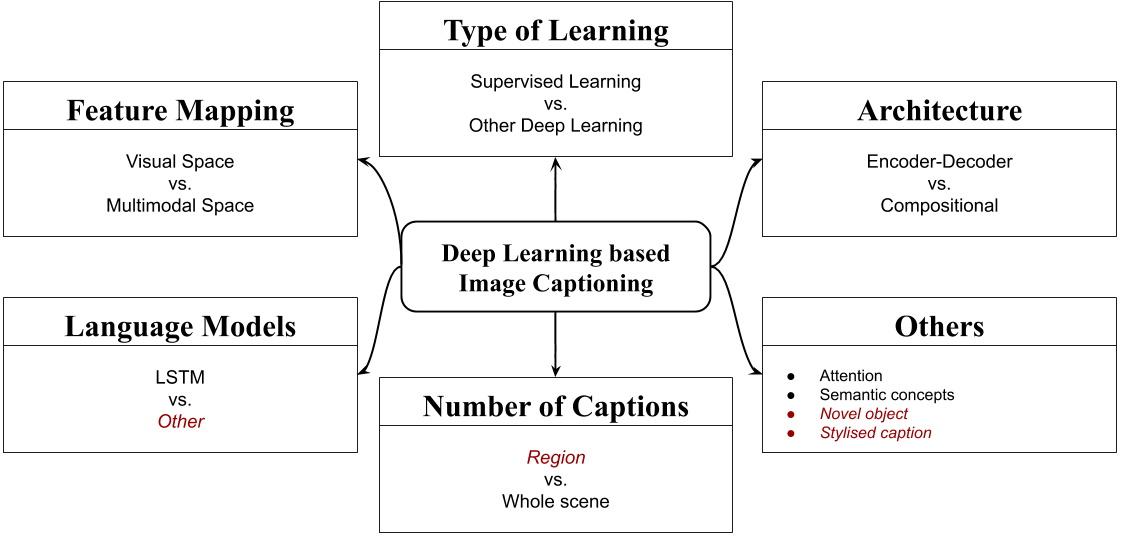} 
    \caption{Aspects of deep learning generic image captioning methods, using the terminology of \citet{Hossain2019}. Aspects that have \emph{not} been used in DC are shown in red and italics.}
    \label{fig:icdc}
\end{figure}

Deep learning approaches are currently dominant in both generic and diagnostic image captioning.\footnote{For a survey of earlier work on generic image-to-text generation, consult \citet{Bernardi2016}.} 
\citet{Hossain2019} compiled a taxonomy of aspects of \emph{generic} image captioning methods that are based on deep learning. Fig.~\ref{fig:icdc} depicts that taxonomy. We highlight aspects that have \emph{not} also been used in \emph{diagnostic} captioning work. The fact that most of the aspects have also been used in DC indicates that  generic image captioning methods are also applicable (or at least have been considered) in DC. The best generic image captioning methods, however, are not necessarily the best ones for DC, mostly because of two factors. First, DC methods do not aim to simply describe what is present in an image, unlike generic image captioning methods. DC aims to report clinically important information that is relevant for diagnostic purposes. Simply reporting, for example, which organs are shown in a medical image is undesirable, if there is nothing clinically important to be reported about them. Second, as we have already discussed in previous sections, diagnostic reports are often very similar across  examinations of different patients. This allows retrieval-based approaches to perform surprisingly well, often challenging encoder-decoder approaches that are currently the state of the art in generic image captioning. We present both approaches and other alternatives below.

Regarding the type of input (`feature mapping' box of Fig.~\ref{fig:icdc}) that is used to generate diagnostic captions, both images \citep{Li2018,Wang2018,Gale2018,Xue2018,Liu2019,Huang2019} and images combined with text \citep{Shin2016b,Jing2018,Yin2019,Yuan2019} have been explored.
Both supervised and reinforcement learning have been employed in DC \citep{Jing2018,Li2018,Liu2019}; the latter falls in the `other' category of learning type in the taxonomy of Fig.~\ref{fig:icdc}. A caption in current DC work typically refers to an entire medical image or even to a set of medical images, for example multiple X-rays from a single examination \citep{Li2018,Yuan2019}, not to a particular region of the image(s), unlike some generic image captioning work \citep{Hossain2019}. The most common system architecture in DC is the encoder-decoder approach. However, other approaches (dubbed `compositional' by \citet{Hossain2019}) have also been tried in DC, for example the knowledge graph approach by \citet{Li2019}. 
Usually, LSTMs (viewed as language models in the taxonomy of Fig.~\ref{fig:icdc}) are employed to generate text in DC, but we observe that Transformer-based models \citep{BERT,GPT2} can also be employed, which is a direction with potential that has just begun being explored \citep{Chen2020}. Regarding the `other' box of Fig.~\ref{fig:icdc}, where miscellaneous other aspects are listed, concepts \citep{Jing2018} and attention-based methods are common in DC \citep{Wang2018,Jing2018,Gale2018,Li2018,Xue2018,Liu2019,Huang2019,Yin2019,Yuan2019}. 
Gaze mechanisms, which have not been explored in generic image captioning \citep{Hossain2019}, could also be useful in DC. For example, attention mechanisms might aim to mimic how a physician focuses in turn on different parts of a medical image.
By contrast, the emotional aspects that have been studied for generic image captioning \citep{Nezami2020}, are not related to DC. `Novel objects' in generic image captioning are objects present only in the test dataset \citep{Agrawal2019}. Novel object captioning has not been investigated in DC yet. However, it would certainly be interesting to assess system-generated text for patients with rare (or new) conditions.
`Stylised captioning', which aims to generate stylised and attractive descriptions of generic images, is of little importance in DC, where informativeness and clinical accuracy matter most.

\subsection*{Diagnostic Captioning Architectures}
Most often, the encoder-decoder architecture \citep{Cho2014} is employed in DC \citep{Bai2018,Liu2019survey}, with or without visual attention and reinforcement learning. Systems that adopt this architecture first encode the medical images as dense vectors, typically using CNN-based image encoders. They then generate the diagnostic text from the image encoding, typically using Recurrent Neural Network (RNN) decoders. 
However, retrieval-based methods have also been proposed for DC \citep{Li2018}, and even their simplest forms (e.g., reusing the report of the visually nearest training instance)  have been found to outperform all other systems in clinical (tag-based) Recall \citep{Liu2019}. As shown in Table~\ref{tab:results}, more elaborate retrieval-based systems can outperform state of the art encoder-decoders in DC. More specifically, the retrieval-based system of \citet{Li2019} achieved overall better results than their earlier hybrid encoder-decoder and retrieval-based approach \citep{Li2018}. The reader is warned that not all of the results of Table~\ref{tab:results} are directly comparable, since some of them are obtained from different datasets, or different training/development/test splits. However, the very high manual evaluation score of the retrieval-based method of \citet{Li2019} is an indication that the encoder-decoder approach may be worse than retrieval-based approaches to DC, and that the latter should be explored more in DC. We also note that retrieval has been recently found to improve language models \citep{Kassner2020,khandelwal2019}. The benefits may be greater when modeling the language of diagnostic reports, where large parts of text are often very similar or exactly the same across different patients, as already discussed.

Having provided a brief overview of the most common DC approaches, we now discuss them in more detail, starting from the encoder-decoder architecture.

\begin{table}
%\begin{adjustwidth}{-1cm}{-1cm}
\begin{threeparttable}
\caption{ Evaluation scores of DC methods, using BLEU-1/-2/-3/-4 (B1, B2, B3, B4), METEOR (M), ROUGE-L (R), CIDEr (C), manual evaluation (ME), clinical F1 (CF1). All scores are percentages. The first column shows the name of the system or the authors who introduced it. The \textsc{From} column shows the article the results were obtained from. For \mimic, we computed CF1 from the micro-averaged Precision and Recall reported by \citet{Liu2019}. The \textsc{Type} column shows the type of method: Retrieval-based (Rb), Encoder-Decoder (ED), or Decoder only (D). The columns VA, HD, RL correspond to visual attention, hierarchical decoding, reinforcement learning, respectively; these are some of the most frequent aspects of ED methods. Datasets marked with $\dagger$ are not publicly available. Scores for the same dataset are strictly comparable only when obtained from the same article (\textsc{From} column), because of different training/development/test splits that may have been used otherwise. The first two systems are baselines. For all other systems, when multiple versions exist, the performance of the most extensive version is reported, as opposed to simpler versions used, for example, in ablation studies.}
\small
\setlength\tabcolsep{3pt}
\begin{tabular}{l c c c c c l c c c c c c c c c}
\hline \sc Sys./Aut. & \sc Type & \sc va &\sc hd & \sc rl & % \tiny
\sc From & \sc Dataset & \sc B1 & \sc B2 & \sc B3 & \sc B4 & \sc M & \sc R & \sc C & \sc ME & \sc CF1 \\ \hline \hline

\multirow{2}{*}{\tiny\bf 1-NN} 
& \multirow{2}{*}{\sc Rb} 
& \multirow{2}{*} 
& \multirow{2}{*}
& \multirow{2}{*} 
& \tiny\citet{Liu2019} & \sc \tiny\iuxray & \bf 23.2 & \bf 11.6 &\bf 5.1 &\bf 1.8 &\bf  &\bf 20.1 &\bf 72.8 & & \\\cline{7-16}
&&&&& \tiny\citet{Liu2019} & \sc \tiny\mimic &\bf 30.5 &\bf 17.1 &\bf 9.8 &\bf 5.7 &\bf  &\bf 24.4 &\bf 75.5 &\bf &\bf 39.1\\ \hline 

\multirow{2}{*}{\tiny\bf BlindRNN} 
& \multirow{2}{*}{\sc D} 
& \multirow{2}{*} 
& \multirow{2}{*} 
& \multirow{2}{*} 
& \tiny\citet{Liu2019} & \sc \tiny\iuxray &\bf 23.3 &\bf 13.0 &\bf 8.7 &\bf 6.1 &\bf  &\bf 29.1 &\bf 74.7 & &\\\cline{7-16}
&&&&& \tiny\citet{Liu2019} & \sc \tiny\mimic &\bf 26.9 &\bf 17.2 &\bf 11.3 &\bf 7.4 &\bf  &\bf 27.2 &\bf 71.6 & &\\ \hline \hline

\bf\tiny\citeauthor{Xue2018} \tiny\citeyear{Xue2018}  & \sc ED & \checkmark & \checkmark & & \tiny\citet{Xue2018} & \sc \tiny\iuxray &\bf 46.4 &\bf 35.8 &\bf 27.0 &\bf 19.5 &\bf 27.4 &\bf 36.6 &  & &\\ \hline

\bf\tiny\citeauthor{Huang2019} \tiny\citeyear{Huang2019}& \sc ED & \checkmark & \checkmark & & \tiny\citet{Huang2019} & \sc \tiny\iuxray &\bf 47.6 &\bf 34.0 &\bf 23.8 &\bf 16.9 &\bf  &\bf 34.7 &\bf 29.7 &  & \\ \hline

\bf\tiny\citeauthor{Shin2016b} \tiny\citeyear{Shin2016b}& \sc ED & & & & \tiny\citet{Shin2016b} & \sc \tiny\iuxray &\bf 78.5 &\bf 14.4 &\bf 4.7 &\bf 0.0 &  &  &  & &\\ \hline

\multirow{3}{*}{\bf\tiny\citeauthor{Wang2018} \tiny\citeyear{Wang2018}} &
\multirow{3}{*}{\sc ED} & 
\multirow{3}{*}{\checkmark} & 
\multirow{3}{*}{} &
\multirow{3}{*}{} &
\tiny\citet{Wang2018} & \sc \tiny Chest X-ray 14$^\dagger$ &\bf 28.6 &\bf 15.9 &\bf 10.3 &\bf 7.3 &\bf 10.7 &\bf 22.6 &  & & \\\cline{7-16}
&&&&& \tiny\citet{Liu2019} & \sc \tiny\iuxray &\bf 33.0 &\bf 19.4 &\bf 12.4 &\bf 8.1 &\bf&\bf 31.1 &\bf 133.4 & & \\\cline{7-16}
&&&&& \tiny\citet{Liu2019} & \sc \tiny\mimic &\bf 33.2 &\bf 21.2 &\bf 14.2 &\bf 9.5 &\bf &\bf 29.6 &\bf 100.4 & & \bf 40.6 \\ \hline

\bf\tiny\citeauthor{Zhang2017b} \tiny\citeyear{Zhang2017b} & \sc ED &\bf \checkmark & & & \tiny\citet{Zhang2017b} & \sc \tiny BCIDR$^\dagger$ &\bf 91.2 &\bf 82.9 &\bf 75.0 &\bf 67.7 &\bf 39.6 &\bf 70.1 &\bf 2.04 & &\\ \hline

\multirow{2}{*}{\bf\tiny \citeauthor{Gale2018} \tiny\citeyear{Gale2018}} &
\multirow{2}{*}{\sc ED} & 
\multirow{2}{*}{\checkmark} & 
\multirow{2}{*}{} & 
\multirow{2}{*}{} & 
\tiny\citet{Gale2018}& \sc \tiny PelvicX$^\dagger$ &\bf 91.9 &\bf 83.8 &\bf 76.1 &\bf 67.7 &  &  &  & & \\ \cline{7-16}
&&&&& \tiny\citet{Gale2018} & \sc \tiny PelvicX T$^\dagger$ &\bf 65.0 &\bf 37.9 &\bf 24.2 &\bf 15.9 &  &  &  & &\\\hline

\multirow{2}{*}{\bf\tiny\citeauthor{Liu2019} \tiny\citeyear{Liu2019}} & \multirow{2}{*}{\sc ED} & 
\multirow{2}{*}{\checkmark} &
\multirow{2}{*}{\checkmark} &
\multirow{2}{*}{\checkmark} & 
\tiny\citet{Liu2019} & \sc \tiny\iuxray &\bf 35.9 &\bf 23.7 &\bf 16.4 &\bf 11.3 &\bf  &\bf 35.4 &\bf 142.4 & & \\\cline{7-16}
&&&&& \tiny\citet{Liu2019} & \sc \tiny\mimic &\bf 31.3 &\bf 20.6 &\bf 14.6 &\bf 10.3 &\bf  &\bf 30.6 &\bf 104.6 &\bf &\bf 33.8 \\\hline

\multirow{2}{*}{\bf\tiny\citeauthor{Jing2018} \tiny\citeyear{Jing2018}} & 
\multirow{2}{*}{\sc ED} & 
\multirow{2}{*}{\checkmark} & 
\multirow{2}{*}{\checkmark} & 
\multirow{2}{*}{} & 
\tiny\citet{Li2019} & \sc \tiny\iuxray &\bf 45.5 &\bf 28.8 &\bf 20.5 &\bf 15.4 &\bf  &\bf 36.9 &\bf 27.7 &\bf 24.1 &\\\cline{7-16}
&&&&& \tiny\citet{Jing2018} & \sc \tiny\peirgross &\bf 30.0 &\bf 21.8 &\bf 16.5 &\bf 11.3 &\bf 14.9 &\bf 27.9 &\bf 32.9 &\bf & \\ \hline

\multirow{2}{*}{\bf\tiny\citeauthor{Li2018} \tiny\citeyear{Li2018}}& 
\multirow{2}{*}{\sc ED/Rb} & 
\multirow{2}{*}{\checkmark} &
\multirow{2}{*}{\checkmark} &
\multirow{2}{*}{\checkmark} &
\tiny\citet{Li2019} & \sc \tiny\iuxray &\bf 43.8 &\bf 29.8 &\bf 20.8 &\bf 15.1 &\bf  &\bf 32.2 &\bf 34.3 &\bf 48.0 & \\\cline{7-16} 
&&&&& \tiny\citet{Li2019} & \sc \tiny CX-CHR$^\dagger$ &\bf 67.3 &\bf 58.7 &\bf 53.0 &\bf 48.6 &\bf  &\bf 61.2 &\bf 289.5 & & \\\hline 

\multirow{2}{*}{\bf\tiny\citeauthor{Li2019} \tiny\citeyear{Li2019}} & \multirow{2}{*}{\sc Rb} & 
\multirow{2}{*}{} &
\multirow{2}{*}{} &
\multirow{2}{*}{\checkmark} &
\tiny\citet{Li2019} & \sc \tiny\iuxray &\bf 48.2 &\bf 32.5 &\bf 22.6 &\bf 16.2 &\bf  &\bf 33.9 &\bf 28.0 &\bf 57.4 &\\\cline{7-16}
&&&&& \tiny\citet{Li2019} & \sc \tiny CX-CHR$^\dagger$ &\bf 67.3 &\bf 58.8 &\bf 53.2 &\bf 47.3 &\bf  &\bf 61.8 &\bf 285.0 &\bf 67.8 &\\\hline

\multirow{2}{*}{\bf\tiny\citeauthor{Vinyals2015} \tiny\citeyear{Vinyals2015}} & 
\multirow{2}{*}{\sc ED} & 
\multirow{2}{*}{} & 
\multirow{2}{*}{} & 
\multirow{2}{*}{} & 
\tiny\citet{Liu2019} & \sc \tiny \iuxray &\bf 26.5 &\bf 15.7 &\bf 10.5 &\bf 07.3 &\bf  &\bf 30.6 &\bf 92.6 & & \\\cline{7-16}
&&&&& \tiny\citet{Liu2019} & \sc \tiny \mimic &\bf 30.7 &\bf 20.1 &\bf 13.7 &\bf 09.3 &\bf  &\bf 30.0 &\bf 88.6 &\bf &\bf 33.1 \\\hline

\multirow{2}{*}{\bf\tiny\citeauthor{Xu2015} \tiny\citeyear{Xu2015}} &
\multirow{2}{*}{\sc ED} & 
\multirow{2}{*}{\checkmark} & 
\multirow{2}{*}{} & 
\multirow{2}{*}{} & 
\tiny\citet{Liu2019} & \sc \tiny\iuxray &\bf 32.8 &\bf 19.5 &\bf 12.3 &\bf 08.0 &\bf  &\bf 31.3 &\bf 127.6 & & \\\cline{7-16}
&&&&& \tiny\citet{Liu2019} & \sc \tiny\mimic &\bf 31.8 &\bf 20.5 &\bf 13.7 &\bf 9.3 &\bf  &\bf 28.8 &\bf 96.7 &\bf &\bf 39.6\\\hline

\bf\tiny\citeauthor{Rennie2017} \tiny\citeyear{Rennie2017}
& \sc ED & \checkmark & & \checkmark 
& \tiny\citet{Li2019} & \sc \tiny\iuxray &\bf 22.4 &\bf 12.9 &\bf 08.9 &\bf 06.8 &\bf  &\bf 30.7 &\bf 29.7 & &\\\hline

\bf\tiny\citeauthor{Lu2017} \tiny\citeyear{Lu2017}& \sc ED & \checkmark & & & \tiny\citet{Li2019} &\sc \tiny\iuxray &\bf 22.0 &\bf 12.7 &\bf 08.9 &\bf 06.9 &\bf  &\bf 30.8 &\bf 29.6 & & \\ \hline

\bf\tiny\citeauthor{Yuan2019} \tiny\citeyear{Yuan2019}& \sc ED & \checkmark & \checkmark & & \tiny\citet{Yuan2019} &\sc \tiny\iuxray &\bf 52.9 &\bf 37.2 &\bf 31.5 &\bf 25.5 &\bf 34.3 &\bf 45.3 & & & \\ \hline

\bf\tiny\citeauthor{Yin2019} \tiny\citeyear{Yin2019}& \sc ED & \checkmark & \checkmark & & \tiny\citet{Yin2019} &\sc \tiny\iuxray &\bf 44.5 &\bf 29.2 &\bf 20.1 &\bf 15.4 &\bf 17.5 &\bf 34.4 &\bf 34.2 & & \\ \hline

\end{tabular}
\label{tab:results} 
\end{threeparttable}
%\end{adjustwidth}
\end{table}

\begin{figure*}
    \centering
    \begin{minipage}{\textwidth}
    \subfigure[Show \& Tell]{
        \centering
        \includegraphics[width=0.45\textwidth]{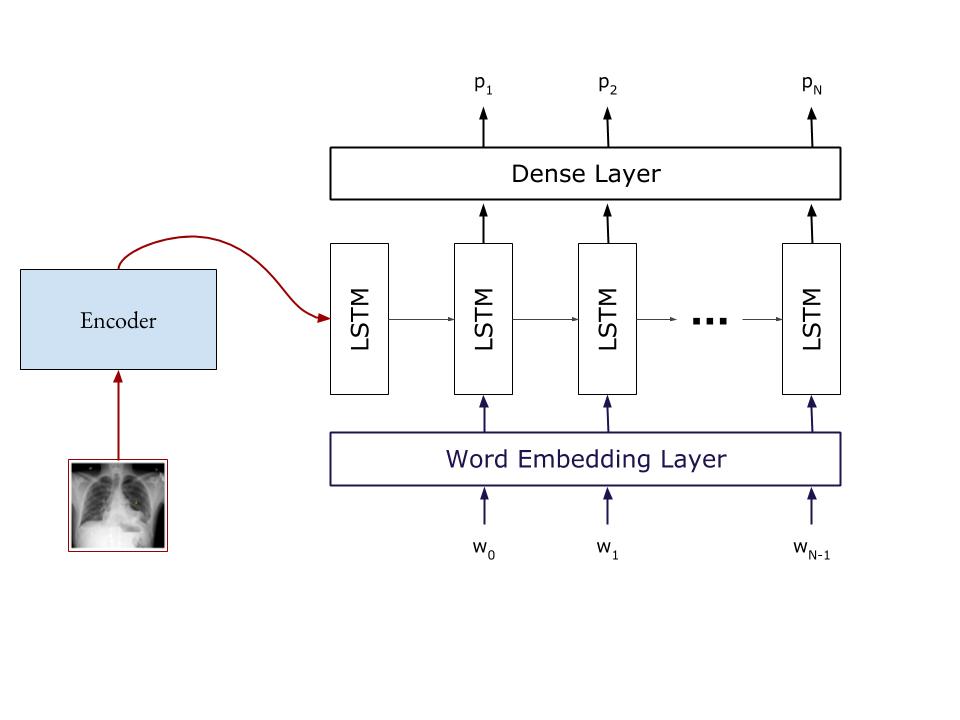}
        \label{fig:vinyals}
    }%\qquad
    ~
    \subfigure[Show, Attend \& Tell]{
        \centering
        \includegraphics[width=0.5\textwidth]{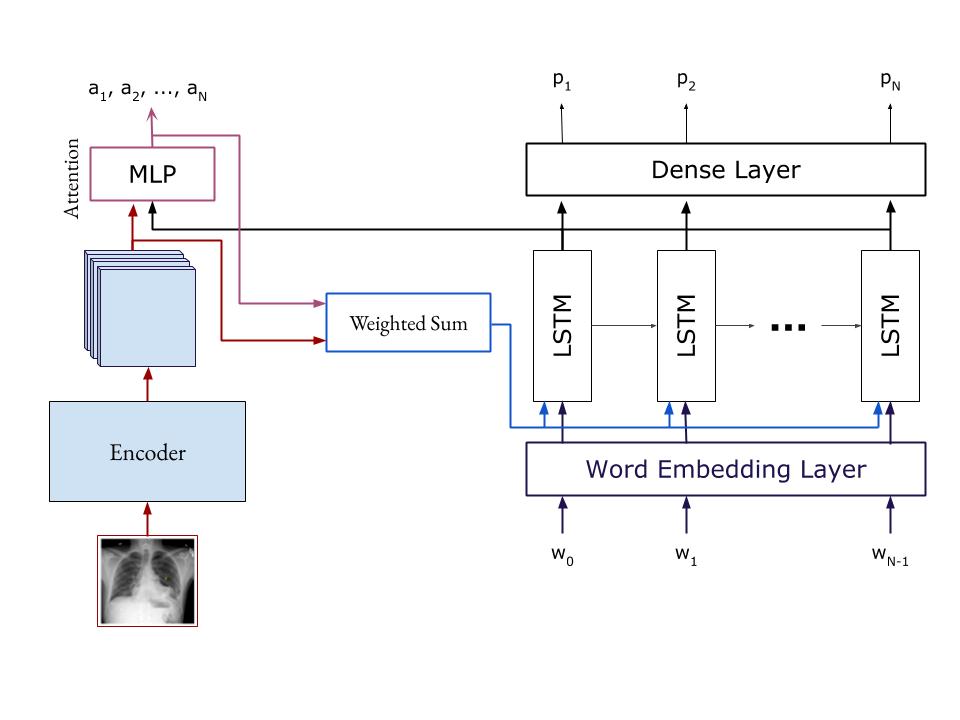}
        \label{fig:xu}
    }%\qquad
    \end{minipage}
    \caption{Left: the Show \& Tell (S\&T) model by \citet{Vinyals2015}. Right: the Show, Attend \& Tell (SA\&T) model by \citet{Xu2015}. S\&T uses the image encoding of the CNN to initialise the LSTM decoder. SA\&T also comprises a visual attention mechanism.}
\end{figure*}

\smallskip
\noindent\textbf{Encoder-Decoder (ED)} The encoder-decoder deep learning architecture was originally introduced for machine translation \citep{Cho2014}, but was then also adopted in generic image captioning \citep{Hossain2019,Bai2018,Liu2019survey}. In machine translation or other text-to-text generation tasks (e.g., summarization) an encoder network, often an RNN such as an LSTM \citep{Hochreiter1997}, reads the input text and converts it to a single vector or a sequence of vectors. A decoder network, often another RNN, then produces the target text, in the simplest case word by word, using as its input the encoding of the input text. An attention mechanism \citep{Xu2015} allows the decoder to focus on particular vectors of the input text encoding, if the latter is an entire sequence of vectors. In generic image captioning, the encoder is typically a CNN \citep{Lecun2015}, which converts the image into a single or multiple vectors (e.g., corresponding to patches of the image). The decoder again produces the target text (caption), using the image encoding as its input. An attention mechanism may again allow the decoder to focus on particular vectors of the image encoding when generating each word; we call mechanisms of this kind `visual attention' and we discuss them separately below. An example of an encoder-decoder model without visual attention for generic image captioning is the model of \citet{Donahue2015}. This model comprises a CNN to encode the image and an LSTM to decode to text. The CNN was CaffeNet \citep{Jia2014} or the better performing VGG \citep{Simonyan2014}. The decoder was an LSTM, which used the representation of the previous generated word along with the image encoding to generate the next word at each timestep. The authors also experimented with a stacked, two-layer LSTM decoder. Another example is the well-known Show \& Tell (S\&T) system of \citet{Vinyals2015}, which was also introduced for generic image captioning; see  Fig.~\ref{fig:vinyals}. It employs the Inceptionv3 CNN \citep{Szegedy2016} to encode the image and uses the image encoding to initialise the LSTM decoder.

\smallskip    
\noindent\textbf{ED + Visual Attention (VA)} We place in this category encoder-decoders that also employ visual attention mechanisms (VA), as in Fig.~\ref{fig:xu} \citep{Xu2015}. Such mechanisms can also be used to highlight on the image the findings described in the report to make the diagnosis more easily interpretable \citep{Zhang2017b,Jing2018,Wang2018,Yin2019,Yuan2019}. 
\citet{Zhang2017b} were the first to employ visual attention  in DC with the MDNet model.\footnote{Zhang et al.\  introduced TandemNet \citep{Zhang2017a} earlier, which also used visual attention, but for medical image classification. TandemNet could also perform captioning, but the authors focused on this task in MDNet.} They used the BCIDR dataset (not publicly available, Section~\ref{sect:data}), which contains pathological bladder cancer images and diagnostic reports, aiming to generate paragraphs conveying findings. MDNet used a form of ResNet \citep{He2016} to encode images. The image encoding acts as the initial hidden state of an LSTM decoder, which also uses visual attention. The decoder was cloned to generate multiple sentences. However, in most evaluation measures the model performed only slightly better than the generic image captioning model of \citet{Karpathy2015} applied to DC.

The system of \citet{Lu2017}, which was designed for generic image captioning and uses visual attention too, was also applied to DC \citep{Li2018}. Its CNN encoder is a ResNet \citep{He2016}, and its decoder is an LSTM. At each timestep, the spatial image encodings (one per image region) and the LSTM hidden state are used as input to a Multi-Layer Perceptron (MLP) with a single hidden layer and a softmax output activation function, acting as a visual attention mechanism \citep{Xu2015}. This mechanism generates one weight per image region, and the weights are used to form an overall weighted image representation. This image representation is then used along with the hidden state of the LSTM decoder to predict the next word. The authors also extended the decoder with a binary gate, which allows deactivating the visual attention when visual information is redundant (e.g., generating stopwords may require no visual attention).

\begin{figure*}
    \centering
    \begin{minipage}{\textwidth}
    \subfigure[Image encoder, visual and semantic features, \\ attention mechanisms, and hierarchical decoder.
    ]{
        \centering
        \includegraphics[width=0.45\textwidth]{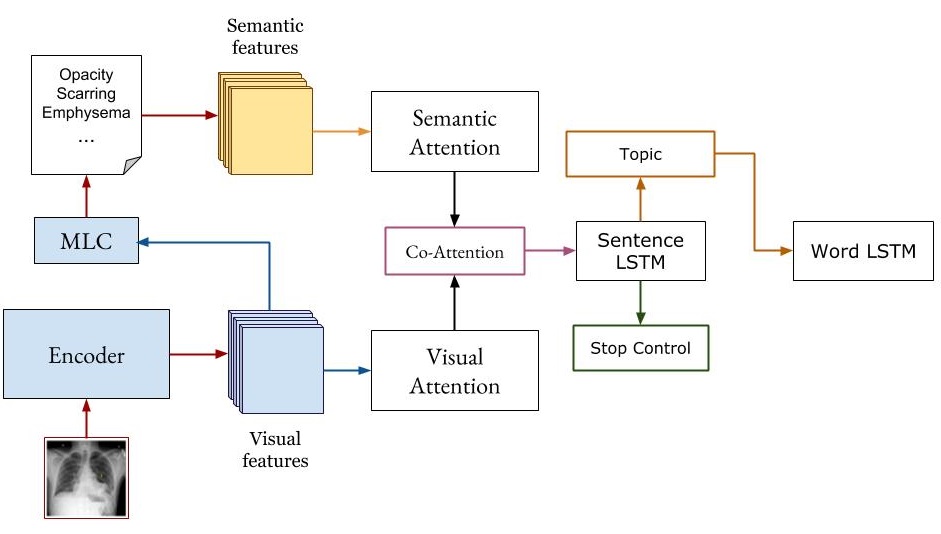}
        \label{fig:jing}
    }%\qquad%
    %~
    \subfigure[Hierarchical text decoder, consisting of a sentence-level LSTM and a word-level LSTM.]{
        \centering
        \includegraphics[width=0.5\textwidth]{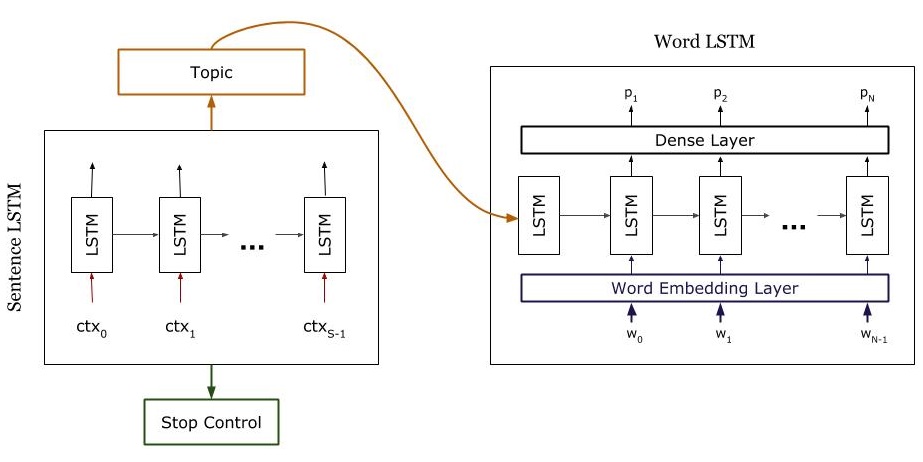}
        \label{fig:hdecode}
    }%\qquad
    \end{minipage}
    \caption{\citet{Jing2018} proposed a DC model that first encodes the image, then extracts `visual' features from the image encoding and `semantic' features from terms predicted from the image encoding. Attention mechanisms are used to produce an overall image representation from the visual and semantic features, which is different at each timestep of the sentence LSTM decoder. At each timestep, the sentence decoder selects the topic of the corresponding sentence. A word LSTM decoder then generates the words of the sentence.}    
\end{figure*}

\citet{Jing2018} created an encoder-decoder model with visual attention especially for DC,
% (see Fig.~\ref{fig:jing}).
illustrated in Fig.~\ref{fig:jing}.
They used VGG-19 \citep{Simonyan2014} to encode each image and extract equally sized patches. Each patch encoding is treated as a `visual' feature vector. An MLP, called MLC in their article and in Fig.~\ref{fig:jing}, is then fed with the visual feature vectors and predicts terms from a pre-determined term vocabulary. The word embeddings (dense vector representations) of the predicted terms of each image are treated as `semantic' feature vectors representing the image. The decoder, which produces the text, is a hierarchical RNN, consisting of a sentence-level LSTM and a word-level LSTM. The sentence-level LSTM produces a sequence of sentence embeddings (vectors), each intuitively specifying the information to be expressed by a sentence of the image description (acting as a sentence topic). For each sentence embedding, the word-level LSTM then produces the words of the corresponding sentence, word by word. More precisely, at each one of its timesteps, the sentence-level LSTM examines both the visual and the semantic feature vectors of the image. An attention mechanism (an MLP fed with the current state of the sentence-level LSTM and each one of the visual feature vectors of the image) assigns attention scores to the visual feature vectors, and the weighted sum of the visual feature vectors (weighted by their attention scores) becomes a visual `context' vector, intuitively specifying which visual features to express by the next sentence. Another attention mechanism (another MLP) assigns attention scores to the semantic feature vectors (representing the image terms), and the weighted sum of the semantic feature vectors (weighted by attention) becomes the semantic context vector, specifying which terms of the image to express by the next sentence. 
At each timestep, the sentence-level LSTM considers the visual and semantic context vectors, produces a sentence embedding (topic), and updates its state, until a stop control instructs it to stop. Given a sentence embedding, the word-level LSTM produces the words of the corresponding sentence, until a special `stop' token is generated. Jing et al.\ showed that their model outperforms generic image captioning models with visual attention \citep{Vinyals2015,Donahue2015,Xu2015,You2016} in DC.
\citet{Wang2018} adopted an approach to DC similar to that of \citet{Jing2018}, using a ResNet-based image encoder and an LSTM decoder, but their LSTM is flat, as opposed to the hierarchical LSTM of \citet{Jing2018}. \citet{Wang2018} also extract additional image features from the states of the LSTM.

\citet{Yin2019} created an encoder-decoder DC model similar to that of \citet{Jing2018}. Again, a hierarchical LSTM attends over image features and representations of abnormality labels predicted from the image encoding. The image encoder is DenseNet \citep{Huang2017}, but Yin et al.\ remove its last global pooling layer (arguing it could lose important spatial information) and the last fully connected layer (which serves as a classifier). Instead they add a convolutional layer that operates on the image region representations and outputs a probability distribution for each particular label over the image, intuitively a heatmap for each label. A global max pooling over each label's heatmap then produces a single probability per label for the entire image. The DenseNet encoder was pre-trained on ImageNet and then fine-tuned on IU X-ray (Section~\ref{sect:data}). The hierarchical LSTM and the attention mechanisms are very similar to those of \citet{Jing2018} discussed above. Yin et al.\ also  add a `topic matching' loss that roughly speaking penalizes topic representations (sentence embeddings) produced by the sentence-level LSTM decoder when they deviate from the representations of the corresponding ground-truth sentences. 

Another encoder-decoder model with visual attention for DC was proposed by \citet{Yuan2019}. It uses the ResNet-152 image encoder \citep{He2016}, which Yuan et al.\ pre-trained on the medical image dataset CheXpert \citep{Irvin2019} to perform multi-label classification with 14 labels (12 disease labels, ``Support Devices" and ``No Finding").\footnote{CheXpert is both the name of a dataset (containing 224,316 chest radiographs) and the name of a tool (`CheXpert labeler') that was used to annotate the dataset. The latter tool was also used in Table~\ref{tab:eval_fail}.} The image encoder was fine-tuned by training it for the same task on IU X-ray (Section~\ref{sect:data}) and then used with a new dense layer added on top for classification with 69 gold labels (medical concepts) extracted from the ground truth reports by SemRep.\footnote{https://semrep.nlm.nih.gov/} Yuan et al.\ show experimentally that an image encoder pre-trained on a large dataset of medical images (224,316 chest X-rays) has better performance than encoders pre-trained on ImageNet. The decoder is again a hierarchical LSTM, in which the sentence LSTM attends over the image and the word LSTM over the medical concepts produced by the encoder. Yuan et al.\ report state-of-the-art results, outperforming \citet{Li2019} and \citet{Jing2018} among others. Yuan et al.\ also allow their model to be fed with multiple images to generate a single report from. This is important, because many imaging examinations comprise multiple images, for example a frontal and a lateral projection image. Systems trained to generate a report from a single medical image at a time cannot handle well such cases. Similar provisions are made in the system of \citet{Li2018}, which also uses Reinforcement Learning, discussed below. 

\smallskip
\noindent\textbf{ED + Reinforcement Learning (RL)} These methods use the encoder-decoder architecture, but also employ Reinforcement Learning (RL) \citep{Sutton2018}. For example, \citet{Rennie2017} employed the REINFORCE algorithm \citep{Williams1992} with a reward based on CIDEr (Section~\ref{ssec:overlap_measure}), but in the context of generic image captioning. An advantage of RL is that non-differentiable evaluation measures can be used directly during training, so that systems are not optimizing loss functions like cross-entropy during training while being assessed with measures such as BLEU, ROUGE, or clinical F1 (Section~\ref{sect:eval}) at test time. For readers not familiar with these issues, we note that when training with backpropagation the loss function must be differentiable, which is not the case for most current DC evaluation measures. By contrast with RL the reward does not need to be differentiable. It can also be given at the end of a sequence of system decisions, in cases where a loss is not available for each individual decision. In the DC system of \citet{Li2018}, for example, RL is used to decide if a sentence will be generated from scratch, or if it will be retrieved from a database with frequently occurring sentences. The image encoding, produced by a DenseNet-121 \citep{Iandola2014} or a VGG-19 \citep{Simonyan2014} CNN, is fed to a hierarchical RNN decoder similar to that of \citet{Jing2018}, illustrated in Fig.~\ref{fig:hdecode}. However, for each sentence embedding (topic) produced by the sentence-level RNN, an agent trained using RL (again using REINFORCE and CIDEr) decides if the sentence will be generated using the word-level RNN or if it will be generated by using a sentence retrieved from a database of frequent sentences. \citet{Li2018} applied their system to \iuxray (Section~\ref{sect:data}), but their experimental results were close to those of a baseline. In more recent work, \citet{Liu2019} used DenseNet-121 \citep{Iandola2014} for image encoding and a hierarchical LSTMs decoder. Similarly to \citet{Li2018} and \citet{Rennie2017}, REINFORCE with a CIDEr-based reward was employed. However, this time RL was used to optimise readability. Liu et al.\ also included a reward based on comparing labels, like the ones of Fig.\ref{fig:captioningb}, extracted by CheXpert \citep{Irvin2019} from the system-generated text and the human-authored report, in order to optimize clinical accuracy.

\smallskip
\noindent\textbf{ED + Language Templates (LT)} 
Template-based generation has a long history in natural language generation \citep{Reiter2000,Gatt2018}, where templates of many different forms have been used, ranging from surface-form sentence templates, to sentence templates at the level of syntax trees, to document structure templates \citep{Deemter2005}. In the context of DC, language templates (LT) have recently been combined with encoder-decoder approaches, attempting to provide more satisfactory diagnostic reports. \citet{Gale2018} focused on classifying hip fractures in frontal pelvic X-rays, and argued that generating reports for such narrow medical tasks can be simplified to using only two sentence templates; one for positive cases, including five placeholders (slots) to be filled in by descriptive terms, and a fixed negative template with no slots. They used DenseNet \citep{Huang2017} to encode the image, and (presumably) classify it as a positive or negative case, and a two-layer LSTM with attention over the image encoding to fill in the slots of the positive template. Their scores are very high (Table~\ref{tab:results}), but this is expected due to the extremely simplified and standardized ground truth reports. For example, the vocabulary of the latter contains only 30 words, including special tokens. 
The DC systems of \citet{Li2018} and \citet{Liu2019} also use sentence templates, but these are rather complete sentences, with no empty slots to fill.

\smallskip
\noindent\textbf{Retrieval-based} approaches to DC can be as simple as reusing the diagnostic text of the visually nearest (in terms of image encoding similarity) medical exam of the training set \citep{Liu2019}. Even this 1-nearest neighbor approach achieves surprisingly good results; see the second row of Table~\ref{tab:results}. Then, it should be no surprise that the more advanced retrieval-based DC approach of \citet{Li2019} outperforms ED \citep{Vinyals2015,Donahue2015}, ED+VA \citep{Jing2018}, and ED+RL \citep{Li2018} methods in Table~\ref{tab:results} (we employ comparable results from the Table). We also note that methods that retrieve sentences \citep{Li2018,Li2019}, discussed above, can also be seen as belonging in the category of retrieval-based systems. Retrieval-based systems were also the top performing submissions of the ImageCLEF Caption Prediction subtask, a task that ran for two consecutive years \citep{ImageCLEF2017,ImageCLEF2018}.\footnote{Later ImageCLEF Caption tasks \citep{ImageCLEFConcept2019,ImageCLEFConcept2020} only required systems to assign labels to medical images, without requiring diagnostic tex to be generated.} The top participating systems of the competition in both years relied on (or included) image retrieval \citep{Liang2017,Zhang2018}. \citet{Zhang2018}, who obtained the best results in 2018, used the Lucene Image Retrieval system (LIRE) to retrieve similar images from the training set, then simply concatenated the captions of the top three retrieved images to obtain the new caption.\footnote{\url{http://www.lire-project.net/}} \citet{Liang2017}, who had the best results in 2017, combined an ED approach with image-based retrieval. They reused a pre-trained VGG encoder and an LSTM decoder, similarly to those of \citet{Karpathy2015}. They trained three such models on different caption lengths and used an SVM classifier to choose the most suitable decoder for the given image. They also used a 1-nearest neighbor method to retrieve the caption of the most similar training image and concatenated it with the generated caption.

\smallskip
\noindent\textbf{Baselines} are included in the first two lines of Table~\ref{tab:results}. BlindRNN is an RNN that simply generates word sequences, having been trained as a language model on medical captions, without considering the image(s); a single-layer LSTM was used in the BlindRNN of Table~\ref{tab:results}. 
The 1-NN baseline retrieves the diagnostic text of the visually most similar image from the training set \citep{Liu2019}. These simplistic baselines were intended to be easy to beat, but as can be seen in Table~\ref{tab:results}, the scores of 1-NN are very high, and they outperform some much more elaborate approaches, such as the system of \citet{Liu2019} in clinical recall.

\section{Conclusions and Directions for Future Research}
\label{sect:conclude}

We have provided an extensive overview of diagnostic captioning (DC) methods, publicly available datasets, and evaluation measures.

In terms of methods, most current DC work uses encoder-decoder deep learning approaches, largely because of their success in generic (non-medical) image captioning. We have pointed out, however, that DC aims to report only information that helps in a medical diagnosis. Prominent objects shown (e.g., body organs) do not need to be mentioned, if there is nothing clinically important to be reported about them, unlike generic image captioning where salient objects (and actions taking place) typically have to be reported. Another major difference from generic image captioning is that medical images vary much less and, consequently, the corresponding diagnostic text is often very similar or even identical across different patients. These two factors allow retrieval-based methods, which reuse diagnostic text from training examples with similar images, to perform surprisingly well in DC. Frequent sentences or sentence templates can also be used, instead of generating them.

For evaluation purposes, DC work has so far relied mostly on word overlap measures, originating from machine translation and summarization, which often fail to capture clinical correctness, as we have also demonstrated using artificial examples. Measures that compare tags (also viewed as labels or classes, corresponding to medical terms or concepts) that are manually or, more often, automatically extracted from system-generated and human-authored diagnostic reports have also been employed, as a means to better capture clinical correctness. They may also fail, however, when the tools that automatically extract the tags are inaccurate, when human annotation guidelines are unclear on exactly which tags should be assigned or not, and when tags cannot fully capture the information to be included in the diagnostic text. Manual evaluation is rare in DC, presumably because of the difficulty and cost of employing evaluators with sufficient medical expertise.

In terms of datasets, we focused on the only two publicly available datasets that are truly representative of the task (\iuxray, \mimic), having first discussed severe shortcomings of the other publicly available datasets (e.g., they may not contain medical images from real examinations). We also collected and reported evaluation results from previous published work for all the DC datasets, methods, and evaluation measures we considered. Although these results are often not directly comparable, because of different datasets or splits used, they provide an overall indication of how well different types of DC methods perform. The results we collected may also help other researchers produce results that will be more directly comparable to previously reported ones. 

\medskip
\noindent Our main findings also guide our proposals for future work on DC, listed below.

\paragraph{Hybrid explainable DC-specific methods.} We believe that hybrid methods, which will combine encoder-decoder approaches that generate diagnostic text from scratch with retrieval-based methods that reuse text from similar past cases are more likely to succeed. Retrieval-based methods often work surprisingly well in DC as already discussed, a fact that DC research has not fully exploited yet, but some editing (or filling in) of recycled previous text (or templates) will presumably still be necessary in many cases, especially when reporting abnormalities. Hence, decoders that tailor (or fill in) previous diagnostic text (or templates) may still be needed. Reinforcement learning can be used to decide when to switch from recycling previous text to editing previous text or to generating new text, as already discussed. Ideally future work will also take into account that physicians do not consider only medical images when diagnosing. They also consider the medical history and profile of the patients (e.g., previous examinations, previous medications, age, sex, occupation). Hence, information of this kind (e.g., from electronic medical records) may need to be encoded, along with the images of the current examination, which may be more than one as already discussed. We also believe that DC methods may also need to become hybrid in the sense that they may need to involve more closely the physician who is responsible for a diagnosis. Current DC work seems to assume that systems should generate a complete diagnostic text on their own, which the responsible physician may then edit. In practice, however, it may be more desirable to allow the physician to see and correct regions of possible interest highlighted by the system on the medical images; then allow the physician to inspect and correct medical terms assigned by the system to the images; then let the physician start authoring the diagnostic text, with the system suggesting sentence completions, re-phrasings, missing sentences, in effect acting as an intelligent image-aware authoring tool. This would allow the physicians to monitor and control more closely the system's predictions and decision making, especially if mechanisms to explain each system prediction or suggestion are available (e.g., highlighting regions on the images that justify each predicted term or suggested sentence completion). 

\paragraph{Better intrinsic and extrinsic evaluation.} We  discussed the shortcomings of current automatic evaluation DC measures. Improving these measures to capture desirable properties of diagnostic text, especially clinical correctness, is hence an obvious area where further research is needed. Advances in evaluation measures for machine translation \citep{Sun2020}, summarization \citep{Xenouleas2019} or, more generally, text generation \citep{Sellam2020} also need to be monitored and ported to DC evaluation when appropriate. Despite the high cost, more manual evaluations of system-generated diagnostic reports by qualified physicians are also needed to obtain a better view of the real-life value of current DC methods and desired improvements. More extrinsic evaluations are also necessary, for example to check if DC methods can indeed shorten the time needed by a physician to author a diagnostic report, if DC methods indeed help inexperienced physicians avoid medical errors, if they reduce the pressure physicians feel etc. Extrinsic evaluations of this kind may also help shift DC methods towards hybrid forms that will involve physicians to a larger extent, as suggested above.  

\paragraph{More, larger, realistic public datasets.} We pointed out that there are currently only two datasets that are truly representative of the DC task. The first one, \iuxray, is rather small (approx.~4k instances) by today's standards. The second one, \mimic, is much larger (approx.~228k instances), but still small compared to the approx.~1~billion imaging examinations performed annually worldwide (Section~\ref{sect:intro}), and it contains only English reports. Hospitals worldwide routinely save diagnostic medical images and the corresponding reports in their systems using established standards, at least for the images and their metadata (e.g., DICOM). Regulations and guidelines to protect sensitive information (e.g., HIPAA) are also available, and automatically removing sensitive information from both images and diagnostic reports seems feasible to a large extent (Section~\ref{sect:data}). Hence, it should be possible to construct many more and much larger publicly available DC datasets in many more languages; ideally these datasets would also include medical records and other information that physicians consult for diagnostic purposes, not just the medical images, as already discussed. What seems to be missing is a set of established, possibly standardized, procedures to construct publicly available and appropriately anonymized DC datasets. In turn, this requires concrete evidence (e.g., from extrinsic evaluations) of the possible benefits that DC methods may bring to public health systems, and well documented best practices.

\section*{Acknowledgements}
\noindent We thank Vasiliki Liakopoulou for language editing of earlier drafts of this article.

\bibliographystyle{apalike}
\bibliography{references}

\end{document}